\documentclass[sigconf]{aamas}

\usepackage{balance} 
\usepackage{multirow}
\usepackage{colortbl}
\definecolor{oursrow}{RGB}{222,235,247}
\usepackage{listings}
\lstdefinestyle{aamas-tight}{
	basicstyle=\ttfamily\small,
	columns=fullflexible,
	numberstyle=\scriptsize\color{gray},
	backgroundcolor=\color{gray!12},
	frame=single,
	framerule=0.4pt,
	rulecolor=\color{black!40},    
	numbers=left,
	numbersep=6pt,
	xleftmargin=8pt,
	columns=fullflexible,
	showstringspaces=false
}
\setcopyright{none}
\acmConference{arXiv preprint}{}{}{}
\renewcommand\footnotetextcopyrightpermission[1]{}
\acmSubmissionID{<<xxxx>>}

\submissionType{Research Paper Track}

\title[LLM-GRCR]{Framing Goal-Directed Reasoning with Critique-Centric Refinement for LLM-Symbolic Planning}

\author{Xuelong Liu}
\affiliation{
	\institution{School of Artificial Intelligence, University of Chinese Academy of Sciences}
	\institution{Institute of Automation, Chinese Academy of Sciences}
	\city{Beijing}
	\country{China}}
\email{liuxuelong2022@ia.ac.cn}

\author{Jiahao Zhao}
\authornote{Corresponding author.}
\affiliation{
	\institution{Institute of Automation, Chinese Academy of Sciences}
    \institution{Wenge AI}
    \institution{ScienceOne AI}
	\city{Beijing}
	\country{China}}
\email{zhaojiahao2019@ia.ac.cn}

\author{Qingchao Kong}
\affiliation{
	\institution{Institute of Automation, Chinese Academy of Sciences}
	\institution{School of Artificial Intelligence, University of Chinese Academy of Sciences}
	\city{Beijing}
	\country{China}}
\email{qingchao.kong@ia.ac.cn}

\author{Daniel Dajun Zeng}
\affiliation{
	\institution{Institute of Automation, Chinese Academy of Sciences}
	\institution{School of Artificial Intelligence, University of Chinese Academy of Sciences}
	\city{Beijing}
	\country{China}}
\email{dajun.zeng@ia.ac.cn}

\author{Wenji Mao}
\authornotemark[1]
\affiliation{
	\institution{Institute of Automation, Chinese Academy of Sciences}
	\institution{School of Artificial Intelligence, University of Chinese Academy of Sciences}
	\city{Beijing}
	\country{China}}
\email{wenji.mao@ia.ac.cn}

\begin{abstract}

	 Planning has been a central field in AI and agent research. Despite recent advances in large language model (LLM) reasoning, accurate planning remains challenging as it requires LLMs to reason over long-horizon action sequence and coordinate multi-step decision toward goal attainment. Symbolic planning, with its explicitly represented action semantics and precisely generated plans, provides a well-structured schema for promoting the planning capabilities of LLMs, represented by LLM-symbolic planning.  However, existing LLM-symbolic planning approaches  typically rely on direct action-sequence generation or free-form reasoning, without explicit mechanisms to frame the representation and causal dependencies underlying plan construction and revision, potentially leading to planning failures. To improve the planning capabilities of LLMs, in this paper,  we propose LLM-GRCR, a structured framework that formulates LLM-symbolic planning as two complementary components—goal-directed plan generation and critique-centric plan refinement—both grounded on a shared structured representation of action semantics, causal dependencies and conflicts. Specifically, goal-directed plan generation utilizes structured representation of causal dependencies to facilitate CoT-like LLM reasoning for reliable plan construction. Critique-centric plan refinement then identifies causal threats and ordering constraints (i.e. critiques) as structured representation to support agentic LLM self-improve. Experiments across eight standard planning domains show that LLM-GRCR achieves the highest average problem-solving rate among the representative baselines, demonstrating the effectiveness of our neuro-symbolic framework with explicitly framing of plan generation and refinement.
	 
\end{abstract}

\newcommand{\BibTeX}{\rm B\kern-.05em{\sc i\kern-.025em b}\kern-.08em\TeX}

\begin{document}
	
	
	
	
    	\maketitle 
	
	\section{Introduction}
	Planning has long been a central research area in AI and agent community that provides the key reasoning and decision-making capabilities for intelligent agents. Recent advances in large language models (LLMs) reasoning have demonstrated their strong capabilities in mathematical reasoning, code generation, and multi-step problem solving~\cite{shao2024deepseekmath, ren2025deepseek, ni2023lever, madaan2023selfrefine, zhao2023large, yao2023tree}. Despite these progresses, LLM planning remains a challenging task, due to its requirement of reasoning over long-horizon action sequence and coordinating multi-step decision toward goal attainment~\cite{pmlr-v235-kambhampati24a}. Symbolic planning~\cite{ghallab2004automated}, with its explicitly represented action semantics, state transitions, goal conditions and precisely generated plans, provides a well-structured representational schema for promoting the planning capabilities of LLMs. Recently, the combination of symbolic planning with LLM reasoning has received increasing research attention, and motivates LLM-symbolic planning as a prominent neuro-symbolic approach~\cite{valmeekam2023planbench}.


	Despite its recent progress, existing LLM-symbolic planning approaches often rely on direct action-sequence generation or free-form chain-of-thought (CoT) reasoning~\cite{verma2025pddlinstruct, fan2026safegen, belcamino2026generalization, rossetti2024learning, pallagani2022plansformer}. Although these approaches can reason about action requirements, they do not always maintain an explicit dependency representation throughout plan construction and correction. Symbolic planning requires coordinating interacting actions under changing conditions, which involves establishing support for goals and preconditions, preserving causal dependencies, and satisfying ordering constraints~\cite{ghallab2004automated}. Failures can arise from unsupported goals or preconditions, causal threats that invalidate required conditions, or conflicting ordering constraints. When these dependencies are not explicitly recorded, an invalid action sequence provides limited guidance about which relationships failed and which decisions require revision.
	
	Structured causal representations from classical planning offer a basis for addressing this challenge~\cite{penberthy1992ucpop, mcallester1991systematic}. By recording which actions support each goal or precondition and the ordering relationships required to preserve that support, they make the dependencies underlying a candidate plan available for inspection. Missing support, causal threats, and ordering conflicts can then be associated with specific actions and dependencies, providing concrete targets for plan diagnosis and revision\cite{pollack1997flaw}. Such representation offers an explicit mechanism as well as a unified structure to frame the representation and causal dependencies underlying plan construction and revision, so as to alleviate potential planning failures.

\begin{figure}[t]
		\centering
		\includegraphics[width=0.85\linewidth]{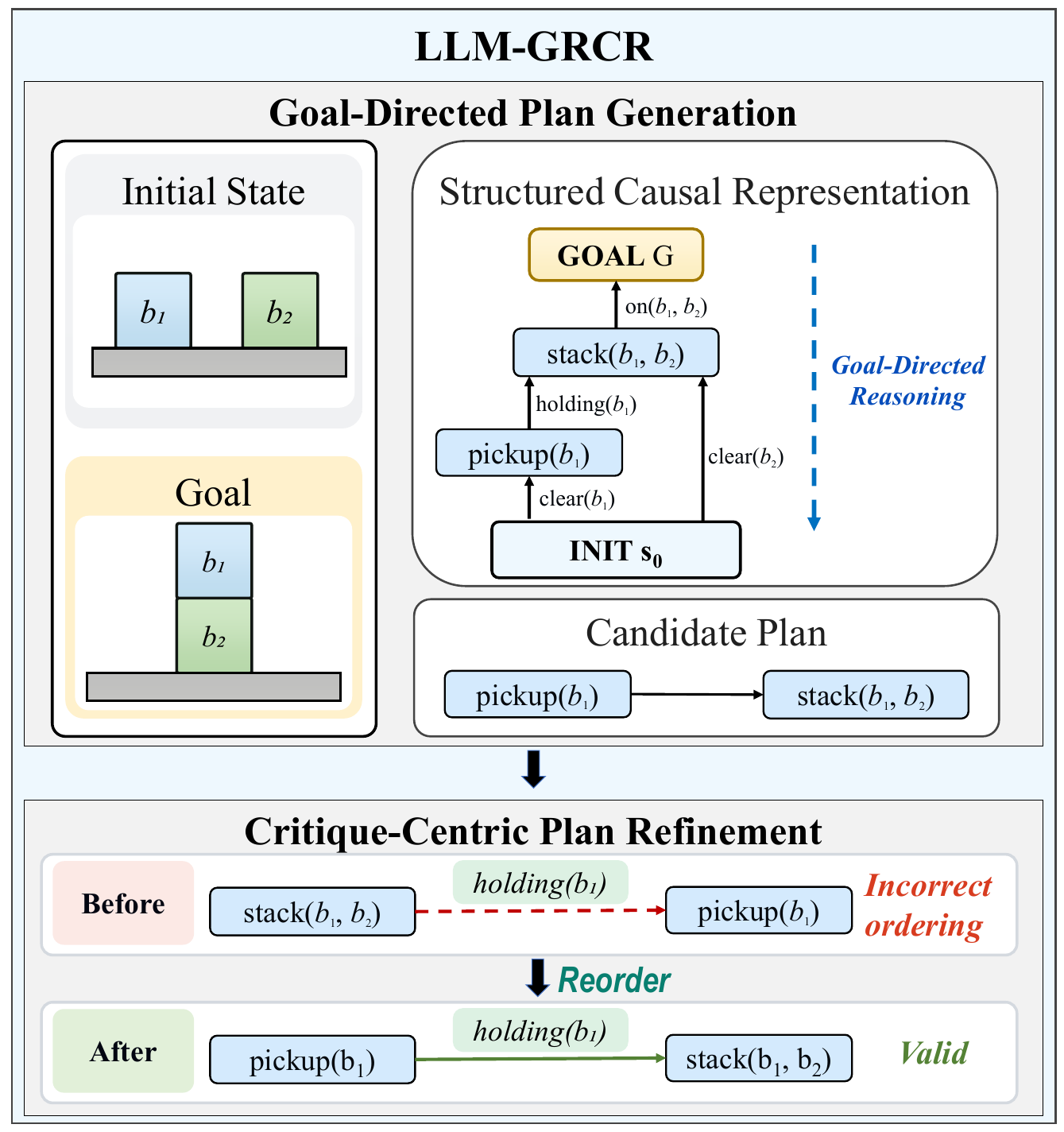}
		\caption{LLM-GRCR illustrated on a Blocksworld problem.}
		\label{fig:Briefmethod}
	\end{figure}

	Motivated by the above considerations, in this paper, we propose  a \textbf{G}oal-directed \textbf{R}easoning and \textbf{C}ritique-centric \textbf{R}efinement (\textbf{LLM-GRCR} ) framework for LLM-symbolic planning. Our framework consists of two components, goal-directed plan generation and critique-centric plan refinement. It trains a goal-directed LLM generator to construct causal dependencies and candidate plans, and an LLM-based revision model to correct diagnosed failures through action-level revisions, through a shared structural causal representation.
	Figure~\ref{fig:Briefmethod} illustrates how the structured causal representation supports both plan generation and refinement. To achieve the goal $\mathrm{on}(b_1,b_2)$, goal-directed reasoning first selects $\mathrm{stack}(b_1,b_2)$ and then traces its required precondition $\mathrm{holding}(b_1)$ backward to the supporting action $\mathrm{pickup}(b_1)$. This yields the candidate plan $\langle \mathrm{pickup}(b_1), \mathrm{stack}(b_1,b_2) \rangle$. During critique-centric refinement, the same representation is used to diagnose an incorrectly ordered plan: $\mathrm{stack}(b_1,b_2)$ fails because its required condition $\mathrm{holding}(b_1)$ is established only by a later $\mathrm{pickup}(b_1)$ action. The revision model can then correct the action order accordingly. In this way, the representation makes explicit both how actions are causally supported during plan construction and how support violations can be identified during execution, thereby connecting goal-directed generation with targeted refinement.


	Our main contributions are summarized as follows:
	\begin{itemize}
		\item We propose LLM-GRCR, a unified neuro-symbolic framework for LLM planning, which connects goal-directed plan generation with critique-centric plan refinement, grounded on the shared structural causal representations.

		\item Our proposed LLM-GRCR frames plan generation and refinement via identifying the representational schema of causal linkages, threats and ordering constraints to facilitate CoT-like  LLM reasoning and agentic LLM self-refinement. 
		
		\item We verify LLM-GRCR across eight representative planning domains, and the results show obvious performance improvements of LLM-GRCR  over baselines, indicating the  effectiveness of our framework for LLM-symbolic planning.
	\end{itemize}

%
	\section{Related Work}

	\subsection{LLMs for Planning}
	
	Recent advances in large language models (LLMs) have demonstrated strong reasoning capabilities\cite{shao2024deepseekmath, ren2025deepseek, ni2023lever}, motivating growing interest in their application to planning and sequential decision making~\cite{zhao2023large, yao2023tree}. Early approaches typically rely on prompting pretrained LLMs to generate plans or action sequences directly from task descriptions, often augmented with reasoning strategies such as chain-of-thought prompting~\cite{wei2022chain, wang2023selfconsistency}. Despite their strong general reasoning ability, however, LLMs remain unreliable on tasks that require long-horizon decision making, precise action dependencies, or strict execution constraints. To improve planning reliability, several approaches incorporate feedback and iterative refinement into the planning process. These approaches allow LLMs to revise intermediate decisions or candidate solutions based on environment observations~\cite{yao2023react}, execution feedback~\cite{shinn2023reflexion}, or self-generated feedback~\cite{madaan2023selfrefine}. Another line of work integrates LLMs with explicit search, using LLMs to propose candidate actions, evaluate intermediate states, or guide exploration over alternative planning trajectories\cite{zhao2023large, yao2023tree}. By combining the generative flexibility of LLMs with systematic search, these approaches can improve planning reliability and reduce dependence on a single generation trajectory.
	
	Despite these advances, most existing approaches improve planning primarily through inference-time mechanisms, often at the cost of additional computational overhead. This has motivated growing interest in directly enhancing LLM planning capabilities through task-specific post-training.

	\subsection{Training LLMs for Symbolic Planning}
	
	Beyond inference-time prompting, another line of research directly adapts LLMs to symbolic planning through task-specific post-training. An early representative approach is Plansformer~\cite{pallagani2022plansformer}, which fine-tunes pretrained Transformer models on planning instances paired with solution plans, enabling direct generation of action sequences from symbolic problem specifications. Subsequent studies~\cite{rossetti2024learning} further explore learning planning policies from collections of solved instances to improve generalization across problems within the same domain.
	
	More recent work introduces explicit reasoning supervision into the training process. PDDL-Instruct~\cite{verma2025pddlinstruct} constructs instruction-tuning data that exposes reasoning about action applicability, state transitions, and plan validity, encouraging language models to acquire planning-specific reasoning patterns rather than merely imitate final action sequences. Other approaches  further combine supervised fine-tuning with reinforcement learning ~\cite{fan2026safegen, belcamino2026generalization}. For example, SafeGen-LLM~\cite{fan2026safegen} performs supervised fine-tuning on constraint-compliant planning data and subsequently applies group-relative policy optimization (GRPO) ~\cite{shao2024deepseekmath} with rewards derived from formal verification to improve constraint satisfaction.
	
	These studies demonstrate the effectiveness of post-training for symbolic planning. However, existing approaches typically supervise action generation, state-transition reasoning, or plan validity, while the structural causal representation among goals, actions, and preconditions remains largely implicit. Our work instead maintains these dependencies as an explicit representation shared across plan generation and refinement. 

	
	
%
	\section{Preliminaries}

	\subsection{Symbolic Planning and PDDL}
	\label{pre:symbolic-planning}
	
	We consider deterministic symbolic planning problems represented in the Planning Domain Definition Language (PDDL). For clarity, we use a STRIPS-style formulation to introduce the planning notation. A PDDL domain is denoted as
	$\mathcal{D}=\langle\mathcal{F},\mathcal{A}\rangle,$
	where $\mathcal{F}$ is a set of predicates describing the symbolic properties of the environment, and $\mathcal{A}$ is a set of action schemas that can be performed in the domain. Each action schema $\alpha\in\mathcal{A}$ is specified by its parameters, preconditions $\operatorname{pre}(\alpha)$, add effects $\operatorname{add}(\alpha)$, and delete effects $\operatorname{del}(\alpha)$. These action definitions characterize how the symbolic state of the environment can change.
	
	Given a domain $\mathcal{D}$, a PDDL problem instance is represented as
	$
		\mathcal{P}=\langle\mathcal{U},s_0,\mathcal{G}\rangle,
	$
	where $\mathcal{U}$ is a finite set of objects used to instantiate the predicates and actions in $\mathcal{D}$, $s_0$ denotes the initial state, and $\mathcal{G}$ denotes the set of goal conditions. The problem instance therefore specifies a concrete planning task under the transition rules defined by the domain.
	
	Grounding the action schemas with the objects in $\mathcal{U}$ produces concrete actions that can be executed in the planning instance.  A state $s$ is represented as a set of ground atoms that hold in the current environment. A grounded action $a$ is applicable in $s$ if
	$
		\operatorname{pre}(a)\subseteq s,
	$
	and its execution produces the successor state
	$
		\gamma(s,a)
		=
		\bigl(s\setminus \operatorname{del}(a)\bigr)\cup \operatorname{add}(a).
	$
	
	A solution to the planning problem is an action sequence
	$
		\pi=\langle a_1,\ldots,a_n\rangle.
	$
	Starting from the initial state $s_0$, the actions in $\pi$ are executed sequentially according to the transition function $\gamma$. The sequence $\pi$ is a valid plan if every action is applicable in the state induced by its preceding actions and the resulting state satisfies all goal conditions,
	$
		\mathcal{G}\subseteq\gamma(s_0,\pi).
	$
	
	Thus, the PDDL domain defines the general action semantics and transition dynamics of the environment, while the PDDL problem instantiates these dynamics with concrete objects, an initial state, and a planning goal. Symbolic planning seeks a valid action sequence that transforms the initial state into a state satisfying the specified goal conditions.
	
	\subsection{Supervised Fine-Tuning and Group-Relative Policy Optimization}
	
	\subsubsection{Supervised fine-tuning.} 
	
	Supervised fine-tuning (SFT) adapts a pretrained language model to a target task by maximizing the likelihood of reference outputs. Given a training set of input-output pairs $\{(x_i,y_i)\}$, the model parameters $\theta$ are optimized by minimizing the negative log-likelihood
	\begin{equation}
		\ell_{\mathrm{SFT}}
		=
		-\sum_i \log p_{\theta}(y_i\mid x_i).
	\end{equation}
	SFT provides a task-specific initialization that encourages the model to produce outputs following the desired reasoning and planning format.
	
	\subsubsection{Group-Relative Policy Optimization.} GRPO ~\cite{shao2024deepseekmath} further improves the policy by comparing multiple candidate responses sampled for the same input. Given a prompt $x$, the current policy samples a group of outputs
	$
		\{y_1,\ldots,y_K\}.
	$
	For each output $y_i$, let $r_i=R(x,y_i)$ denote its reward. In standard GRPO, the corresponding group-relative advantage is computed as
	\begin{equation}
		\hat A_i
		=
		\frac{
			r_i-\operatorname{mean}(r_1,\ldots,r_K)
		}{
			\operatorname{std}(r_1,\ldots,r_K)+\epsilon
		}.
	\end{equation}
	This relative advantage provides a learning signal that favors higher-reward outputs within the same group. In symbolic planning, such rewards can be computed using deterministic criteria derived from PDDL semantics and plan verification.
	

%
\section{Methodology}
\begin{figure*}[t]
	\centering
	\includegraphics[width=0.99\linewidth]{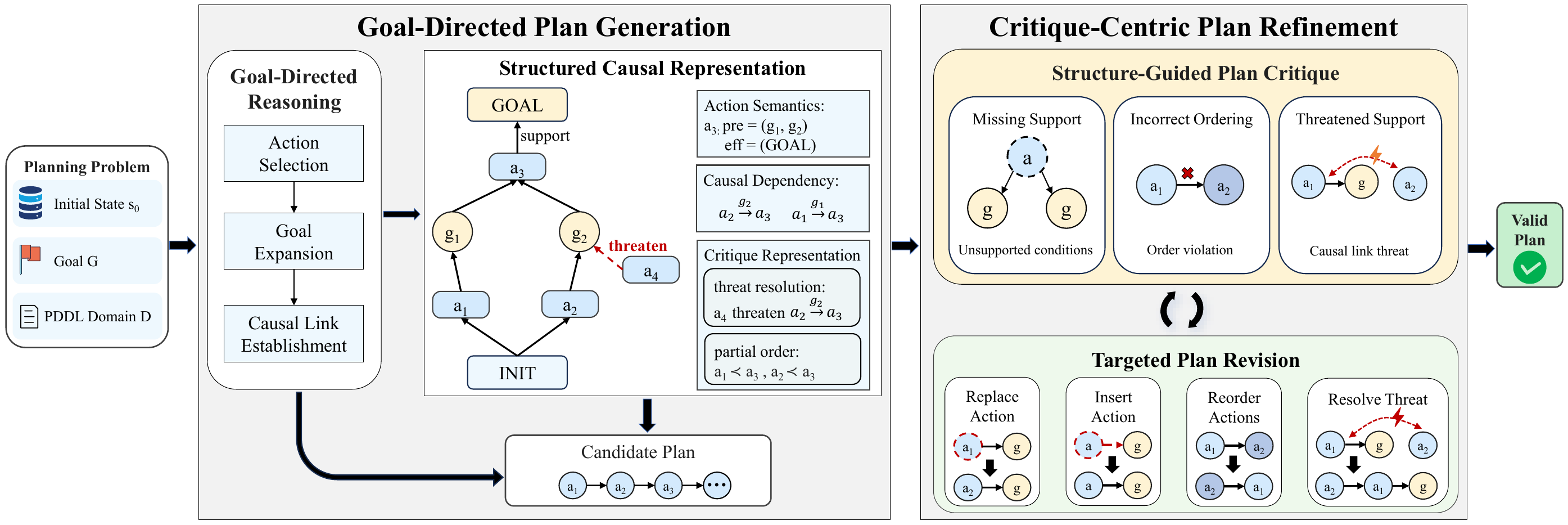}
	\caption{Workflow of LLM-GRCR. Goal-directed generation constructs causal dependencies and a candidate plan, while structure-guided critique provides localized feedback for iterative targeted revision. Here, $a_i$ denotes an action and $g_i$ denotes an intermediate condition.}
	\label{fig:Detailmethod}
\end{figure*}

In this section, we introduce LLM-GRCR, a two-stage framework that unifies goal-directed plan generation and critique-centric plan refinement through a shared structured causal representation, as illustrated in Figure~\ref{fig:Detailmethod}. (1) Goal-Directed Plan Generation: A trained LLM generator selects actions to achieve unresolved goals, recursively expands unsatisfied preconditions into subgoals, and tracks causal dependencies to construct a candidate plan. (2) Critique-Centric Plan Refinement: A symbolic critique module evaluates the candidate plan by examining causal support, ordering constraints, and threats to identify planning errors. Based on this feedback, a trained LLM revision model performs targeted action insertion, replacement, deletion, or reordering while preserving unaffected plan segments whenever possible.

\subsection{Structured Causal Representation}

Building on the causal-link view of partial-order planning~\cite{penberthy1992ucpop, mcallester1991systematic}, LLM-GRCR organizes planning information for a domain--problem pair $(\mathcal{D},\mathcal{P})$ into a structured causal representation
\begin{equation}
	\mathcal{Z}=\langle\mathcal{A}_{\pi},\mathcal{L},\mathcal{O},\mathcal{T}\rangle,
\end{equation}
where $\mathcal{A}_{\pi}$ is the set of grounded actions in the candidate plan $\pi$, $\mathcal{L}$ the set of causal links, $\mathcal{O}$ the set of ordering constraints, and $\mathcal{T}$ the set of potential causal threats. Together, these elements provide three complementary views of the planning process. The \emph{action semantic representation} $\mathcal{A}_{\pi}$ captures the semantics of individual action instances, the \emph{causal dependency representation} $\mathcal{L}$ describes how conditions are established and consumed through causal links. Building on the causal dependency representation $\mathcal{L}$ and the associated action semantics $\mathcal{A}_{\pi}$, the \emph{critique representation} $(\mathcal{O},\mathcal{T})$ captures ordering constraints required to preserve causal support and potential threats that may invalidate these dependencies.



\subsubsection{Action Semantic Representation.}
Each action $a_i\in\mathcal{A}_{\pi}$ in the candidate plan is characterized by its preconditions and effects, which provide the semantic basis for identifying the conditions it requires, establishes, or invalidates.

\subsubsection{Causal Dependency Representation.}
A causal link
$
	a_i\xrightarrow{p}a_j
$
captures a producer--condition--consumer dependency, where the producer $a_i$ establishes a literal $p$ required by the consumer $a_j$. Each causal link therefore induces the ordering relation $a_i\prec a_j$. To uniformly represent dependencies involving the planning boundaries, we use $\mathrm{INIT}$ and $\mathrm{GOAL}$ as special endpoints, such that $\mathrm{INIT}\xrightarrow{p}a_j$ denotes support from the initial state and $a_i\xrightarrow{p}\mathrm{GOAL}$ denotes support for a goal condition. Collectively, these links organize actions around the conditions they establish and consume, making explicit the dependency structure that connects the initial state to the planning goal. This structure further supports goal-directed reasoning by tracing unresolved conditions backward to candidate actions that can establish them.


\subsubsection{Critique Representation.}
The critique representation captures ordering constraints and potential threats associated with causal dependencies $\mathcal{L}$. Specifically, $\mathcal{O}$ records ordering constraints required to preserve causal support, while $\mathcal{T}$ captures potential threats to established causal links. An action $a_k$ threatens a causal link $a_i\xrightarrow{p}a_j$ if it can invalidate $p$ when placed between the producer and consumer. Such threats can be resolved by imposing either
$
a_k\prec a_i
\quad\text{or}\quad
a_j\prec a_k,
$
provided that the resulting ordering remains consistent. Together, $\mathcal{O}$ and $\mathcal{T}$ characterize the structural constraints and potential conflicts associated with causal dependencies  $\mathcal{L}$.




The structured causal representation $\mathcal{Z}$ provides a unified structural foundation for both plan generation and refinement. During generation, it supports the construction of causal dependencies and candidate plans, while during refinement, it enables structure-guided failure diagnosis and targeted plan repair.

%

\subsection{Goal-Directed Plan Generation}

Given a PDDL domain--problem pair $(\mathcal{D},\mathcal{P})$, the trained LLM generator constructs a candidate plan through \textit{goal-directed reasoning} over the \textit{structured causal representation} introduced above. Rather than directly generating an action sequence from the initial state toward the goal, the model reasons backward from unresolved goal conditions and progressively identifies the actions and supporting conditions required for their achievement. For each target condition $p$, the model selects a supporting action $a_j$ whose effects can establish $p$. When $p$ is required by a downstream action $a_i$, the model establishes a causal link $a_j\xrightarrow{p}a_i$, thereby recording their causal dependency. The unsatisfied preconditions of $a_j$ are then introduced as new subgoals and recursively resolved in the same manner. A subgoal is considered resolved if it is satisfied by the initial state or supported by an already introduced action, provided that the corresponding condition remains valid until it is required. Through repeated \emph{action selection}, \emph{subgoal expansion}, and\emph{causal-link establishment}, the LLM incrementally constructs a causal dependency representation $\mathcal{L}$ that connects selected actions through their supporting conditions and captures how goal attainment depends on intermediate actions.


Guided by the action semantic representation $\mathcal{A}_{\pi}$ and the constructed causal dependencies $\mathcal{L}$, the LLM organizes the selected actions into a candidate plan
$
\pi=\langle a_1,\ldots,a_n\rangle,
$
aiming to place supporting actions before their dependent actions.

%
\subsection{Critique-Centric Plan Refinement}

Let $\pi^{(0)}$ denote the candidate plan produced by the LLM generator. LLM-GRCR refines an invalid candidate through an iterative critique-and-revision process. At each round, a symbolic critique module identifies and localizes planning failures within the dependency structure, after which an LLM-based revision policy revises the affected portion of the plan according to the critique while preserving valid decisions whenever possible.

\subsubsection{Structure-Guided Plan Critique}
Given a candidate plan $\pi^{(k)}$, the symbolic critique module first employs VAL~\cite{howey2004val} to verify plan validity by replaying the action sequence from the initial state and checking goal satisfaction. If execution fails, the module identifies the unsatisfied preconditions of the failing action; if execution completes without achieving the goal, it identifies the remaining unsatisfied goal conditions. 
Based on an analysis of failures in generated plans, we identify three categories of causal planning failures. Using the action semantic representation$\mathcal{A}_{\pi}$, causal dependency representation $\mathcal{L}$, and critique representation $(\mathcal{O},\mathcal{T})$, the module diagnoses these failures as follows:
\begin{itemize}
	\item \textbf{Missing support}: A required precondition or goal condition lacks a supporting action or initial-state support.
	\item \textbf{Incorrect ordering}: A supporting action exists but is placed after the action requiring its effect.
	\item \textbf{Threatened support}: An established condition is invalidated before it is required by a consumer or evaluated as a final goal.
\end{itemize}
For each detected failure, the module identifies the affected condition, the corresponding action or goal, and relevant supporting or threatening actions when applicable. This localized causal feedback $\delta^{(k)}$ guides the trained LLM revision model in repairing the affected portions of the plan.

\subsubsection{Critique-Centric Revision}

Given the localized causal feedback $\delta^{(k)}$, the trained LLM revision model focuses on the affected portion of the candidate plan. Its input includes the original planning specification, the current candidate plan $\pi^{(k)}$, and the causal feedback provided by the symbolic critique module. Based on this context, the model proposes local action revisions, such as inserting, deleting, replacing, or reordering actions, to address the diagnosed failure while preserving unaffected portions of the plan whenever possible.

Formally, the revision step is defined as
\begin{equation}
	\pi^{(k+1)}
	=
	\mathrm{Apply}\!\left(
	\pi^{(k)},
	f_{\phi}
	(\mathcal{D},\mathcal{P},\pi^{(k)},\delta^{(k)})
	\right),
\end{equation}
where $f_{\phi}$ denotes the trained LLM revision model with parameters $\phi$, which predicts targeted local revisions to the candidate action sequence. The revised plan is then checked for executability and goal satisfaction. If it remains invalid, the critique module analyzes the revised plan and provides new localized causal feedback for the next revision round. The process continues until the plan is validated or the predefined refinement budget is exhausted.

\begin{table*}[t]
	\centering
	\caption{Problem-solving rate (\%) across eight planning domains. }
	\label{tab:main}
	\small
	\setlength{\tabcolsep}{3pt}
	\begin{tabular}{lrrrrrrrrr}
		\toprule
		Method / setting & Blocksworld & Briefcaseworld & Ferry & Fridge & Grippers & Logistics & Miconic & Satellite & Average \\
		\midrule
		\multicolumn{10}{l}{\emph{External LLMs}} \\
		MiniMax-M2.5 & 5.00 & 18.33 & 30.00 & 30.00 & 13.33 & 8.33 & 8.33 & 0.00 & 14.17 \\
		GLM-5 & 20.00 & 66.67 & 85.00 & 85.00 & 71.67 & 38.33 & \textbf{98.33} & 45.00 & 63.75 \\
		GPT-5-mini & 70.00 & \textbf{96.67} & \textbf{100.00} & 81.67 & \textbf{100.00} & 68.33 & 95.00 & 48.33 & 82.50 \\
		Qwen3.5-Plus & \textbf{95.00} & 95.00 & 98.33 & 90.00 & 93.33 & 80.00 & 96.67 & 56.67 & 88.12 \\
		Kimi-K2.5 & 86.67 & \textbf{96.67} & \textbf{100.00} & 91.67 & 93.33 & \textbf{83.33} & 96.67 & 66.67 & 89.38 \\
		\midrule
		\multicolumn{10}{l}{\emph{Qwen3.5-9B prompting}} \\
		Direct & 1.67 & 0.00 & 0.00 & 0.00 & 0.00 & 0.00 & 0.00 & 0.00 & 0.21 \\
		CoT & 18.33 & 38.33 & 80.00 & 23.33 & 55.00 & 16.67 & 31.67 & 0.00 & 32.92 \\
		Self-refine & 13.33 & 40.00 & 81.67 & 23.33 & 58.33 & 23.33 & 35.00 & 3.33 & 34.79 \\
		Self-consistency CoT & 26.67 & 50.00 & 96.67 & 38.33 & 73.33 & 35.00 & 33.33 & 13.33 & 45.83 \\
		\midrule
		\multicolumn{10}{l}{\emph{Qwen3.5-9B training}} \\
		PDDL-Instruct & 86.67 & 93.33 & 35.00 & 48.33 & 73.33 & 65.00 & 13.33 & 20.00 & 54.38 \\
		Plan-SFT & 33.33 & 90.00 & 98.33 & 93.33 & 83.33 & 68.33 & 81.67 & 61.67 & 76.25 \\
		\rowcolor{oursrow}
		LLM-GRCR & 83.33 & 95.00 & 96.67 & \textbf{98.33} & 95.00 & 80.00 & 81.67 & \textbf{96.67} & \textbf{90.83} \\
		\bottomrule
	\end{tabular}
\end{table*}
\begin{table}[t]
	\centering
	\caption{Problem-solving rate (\%) on easy, medium, and hard planning instances. }
	\label{tab:main-difficulty}
	\setlength{\tabcolsep}{4.5pt}
	\small
	\renewcommand{\arraystretch}{0.95}
	\begin{tabular}{lcccc}
		\toprule
		Method & Easy & Medium & Hard & Average \\
		\midrule
		\multicolumn{5}{l}{\emph{External LLMs}} \\
		MiniMax-M2.5      & 28.75 & 6.88  & 6.88  & 14.17 \\
		GLM-5            & 71.88 & 61.25 & 58.12 & 63.75 \\
		GPT-5-mini &95.00 &82.50&70.00&82.50 \\
		Qwen3.5-Plus    & 98.12 & 88.75 & 77.50 & 88.12 \\
		Kimi-K2.5        & \textbf{99.38} & 90.00 & 78.75 & 89.38 \\
		\midrule
		\multicolumn{5}{l}{\emph{Qwen3.5-9B prompting}} \\
		Direct                     & 0.62  & 0.00  & 0.00  & 0.21 \\
		CoT     & 51.88 & 27.50 & 19.38 & 32.92 \\
		Self-refine                & 56.25 & 26.88 & 21.25 & 34.79 \\
		Self-consistency CoT       & 72.50 & 38.12 & 26.88 & 45.83 \\
		\midrule
		\multicolumn{5}{l}{\emph{Qwen3.5-9B training}} \\
		PDDL-Instruct              & 87.50 & 54.37 & 21.25 & 54.38 \\
		Plan-SFT &92.50&79.37&56.87&76.25\\
		\rowcolor{oursrow}
		LLM-GRCR                & 96.25 & \textbf{93.75} & \textbf{82.50} & \textbf{90.83} \\
		\bottomrule
	\end{tabular}
\end{table}
%
\begin{table*}[!htbp]
	\centering
	\caption{Ablation of generation training and critique-guided refinement in LLM-GRCR. }
	\label{tab:component-ablation-v2}
	\small
	\setlength{\tabcolsep}{3pt}
	\begin{tabular}{lrrrrrrrrr}
		\toprule
		Method / setting & Blocksworld & Briefcaseworld & Ferry & Fridge & Grippers & Logistics & Miconic & Satellite & Average \\
		\midrule
		\rowcolor{oursrow}
		LLM-GRCR 
		& \textbf{83.33} & \textbf{95.00} & 96.67 & \textbf{98.33} & 95.00 & 80.00 & \textbf{81.67} & \textbf{96.67} & \textbf{90.83} \\
		
		w/o refinement
		& 75.00 & 86.67 & 95.00 & 96.67 & 81.67 & 61.67 & 61.67 & 95.00 & 81.67 \\
		
		w/o GRPO 
		& 55.00 & 90.00 & \textbf{100.00} & 96.67 & \textbf{96.67} & \textbf{83.33} & 80.00 & 93.33 & 86.88 \\
		
		w/o GRPO \& refinement
		& 43.33 & 83.33 & 98.33 & 96.67 & 83.33 & 66.67 & 68.33 & 91.67 & 78.96 \\
		
		w/o causal representation \& refinement & 38.33 & 95.00 & 98.33 & 90.00 & 88.33 & 80.00 & 81.67 & 63.33 & 79.38 \\
		\bottomrule
	\end{tabular}
\end{table*}
%

\begin{table}[!htbp]
	\centering
	\caption{Effect of the structural reward. Results are reported as PSR (\%) across eight planning domains.}
	\label{tab:reward-config}
	\small
	\setlength{\tabcolsep}{6pt}
		\renewcommand{\arraystretch}{0.95}
	\begin{tabular}{lrr}
		\toprule
		Domain 
		& w/o $R_{\mathrm{struct}}$
		& with $R_{\mathrm{struct}}$ \\
		\midrule
		Blocksworld    
		& 78.33 & \textbf{83.33} \\
		
		Briefcaseworld 
		& \textbf{96.67} & 95.00 \\
		
		Ferry          
		& \textbf{98.33} & 96.67 \\
		
		Fridge         
		& \textbf{98.33} & \textbf{98.33} \\
		
		Grippers       
		& 86.67 & \textbf{95.00} \\
		
		Logistics      
		& \textbf{83.33} & 80.00 \\
		
		Miconic        
		& \textbf{83.33} & 81.67 \\
		
		Satellite      
		& 90.00 & \textbf{96.67} \\
		\midrule
		Average        
		& 89.37 & \textbf{90.83} \\
		\bottomrule
	\end{tabular}
\end{table}%

%
\subsection{Multi-Stage Training for LLM-GRCR}
LLM-GRCR employs a multi-stage training strategy for goal-directed plan generation and critique-centric refinement. The LLM generator is first initialized with supervised goal-directed planning trajectories and then optimized with GRPO using verifier-grounded rewards. The LLM revision model is trained separately on localized planning failures paired with targeted revisions. In this way, the LLM generator learns to construct causal dependency structures together with their corresponding candidate plans, while the revision model specializes in correcting localized violations identified by the symbolic critique module.

\subsubsection{Training for Goal-Directed Plan Generation}
\paragraph{Structured Planning Trajectory Construction.}
To supervise goal-directed plan generation, we construct structured planning trajectories from valid reference plans. For each PDDL instance, a reference plan is grounded and replayed to recover the support relations underlying its execution. Starting from the goal conditions, the construction process traces supporting actions backward and recursively expands their required conditions, following the goal-directed reasoning procedure described above. This yields a goal-directed causal dependency trace aligned with the corresponding action sequence.

Each trajectory is paired with the original PDDL specification and retained only when both the action sequence and its causal support relations are consistent with the action semantics. The resulting data are used for supervised fine-tuning, providing the LLM generator with an initialization for subsequent policy optimization.

\paragraph{Causal-Representation-Guided Policy Optimization.}
Starting from the supervised generator, we further optimize plan generation with GRPO. For each planning instance, multiple candidate responses are sampled and evaluated using three complementary reward signals:
\begin{equation}
	R
	=
	R_{\mathrm{fmt}}
	+
	R_{\mathrm{val}}
	+
	R_{\mathrm{struct}}.
\end{equation}
The \emph{format reward} $R_{\mathrm{fmt}}$ encourages outputs that follow the required generation protocol and conform to PDDL action syntax, valid action names, and the expected number of action parameters.  The \emph{plan-validity reward} $R_{\mathrm{val}}$ evaluates whether the generated action sequence is executable and achieves the specified goal, while also rewarding partial progress toward goal satisfaction. The \emph{structural reward} $R_{\mathrm{struct}}$ encourages consistent causal links with adequate coverage, consistent condition updates, and acyclic dependencies. The computable reward definitions and configured weights are provided in Appendix E.

Together, these rewards encourage the LLM generator to produce higher-quality plans by leveraging structured reasoning to guide plan generation. In particular, plan validity provides an objective task-level signal, whereas the structural reward encourages semantically consistent causal reasoning to support the generation of more reliable plans.

\subsubsection{Training for Critique-Centric Plan Refinement}



The LLM revision model is trained on localized planning failures paired with targeted plan revisions. Guided by the failure patterns identified above, we construct invalid candidates from verified plans by introducing corresponding structural violations. Specifically, these violations are introduced by removing actions that provide causal support, perturbing producer--consumer orderings, or inserting actions that threaten established causal links. Rather than directly using the predefined corruption types as diagnostic labels, each corrupted plan is re-evaluated by the symbolic critique module, and only examples whose diagnosed failures match the intended violations are retained.

For each retained case, the symbolic critique module provides the localized causal feedback, including the affected condition, its consumer, the relevant producer, and any threatening action when applicable. The corresponding verified plan is then used to derive the local corrections required to resolve the diagnosed violation, rather than serving as a target for complete plan regeneration. The resulting supervision maps localized causal feedback to revisions of the candidate action sequence, while encouraging preservation of unaffected actions. 

The revision model $f_{\phi}$ is trained to map a planning instance, an invalid candidate plan, and its localized causal feedback to targeted action revisions:
\begin{equation}
	f_{\phi}:
	(\mathcal{D},\mathcal{P},\pi,\delta)
	\rightarrow
	\Delta\pi,
\end{equation}
where $\delta$ denotes the localized causal feedback and $\Delta\pi$ denotes targeted revisions to the candidate action sequence for correcting the diagnosed failure. This specialization enables the revision model to translate structure-aware causal feedback into targeted action revisions without regenerating the complete plan from scratch. 

\section{Experiments}
\label{sec:experiments}

We evaluate LLM-GRCR from three perspectives: overall planning performance, the contributions of goal-directed generation and critique-guided refinement, and the effects of key design choices in training and revision.

\subsection{Experimental Setup}
\label{sec:experimental-setup}

\subsubsection{Benchmark.}
 Prior LLM-based planning approaches~\cite{verma2025pddlinstruct,pallagani2022plansformer} often just focus on a few STRIPS domains. In contrast, we evaluate our framework on eight PDDL planning domains from the International Planning Competition (IPC) benchmarks: Blocksworld, Briefcaseworld, Ferry, Fridge, Grippers, Logistics, Miconic, and Satellite. The selected domains span diverse deterministic classical planning problems, covering both standard STRIPS and more expressive PDDL features, including typing, negative preconditions, quantified conditions, and conditional effects. For each domain, problem instances are generated using existing PDDL problem generators~\cite{seipp2022pddl}. We then use Fast Downward~\cite{helmert2006fast} to obtain reference plans and validate the resulting plans with VAL~\cite{howey2004val} to ensure their correctness. The validated instances are then categorized by reference-plan length into three difficulty levels: easy, medium, and hard, corresponding approximately to plans with up to 40 actions, 40--80 actions, and more than 80 actions, respectively. We sample 20 instances from each difficulty level, yielding 60 problems per domain and 480 problems in total. Training and test instances are disjoint at the problem-instance level, verified through hash-based duplicate checks.

\subsubsection{Baselines.}
We compare LLM-GRCR with three groups of baselines. Prompting-based methods include direct generation,   self-consistency CoT~\cite{wang2023selfconsistency}, CoT~\cite{wei2022chain}, and self-refine~\cite{madaan2023selfrefine}. Self-consistency CoT and self-refine each use three sampled outputs. Training-based baselines include Plan-SFT, which directly supervises plan generation without structured reasoning, and a same-backbone reproduction of PDDL-Instruct with three refinement rounds~\cite{verma2025pddlinstruct}. All training-based methods use the same set of 24,000 training examples, comprising 3,000 examples from each domain, evenly distributed across the easy, medium, and hard groups.  All prompting- and training-based methods use the same Qwen3.5-9B backbone. We additionally report results from external LLMs for reference. 

\subsubsection{Implementation and Reporting.}
All controlled experiments use Qwen3.5-9B as the backbone. The LLM-GRCR framework includes supervised training for goal-directed generation, GRPO optimization with the three reward components described above, and up to three rounds of critique-centric refinement. Detailed training and inference hyperparameters are provided in Appendix C. The code is also provided in the supplementary materials.

\subsubsection{Metrics.}
Our primary metric is problem-solving rate (PSR). For domain $d$, let $\mathcal{I}_d$ be its set of test-instance indices, with $N_d=|\mathcal{I}_d|=60$. Then
\begin{equation}
	\mathrm{PSR}_d
	=\frac{1}{N_d}
	\sum_{i\in\mathcal{I}_d}
	\mathbf{1}[\mathrm{VAL}(\hat{\pi}_i)=1].
	\label{eq:psr-v2}
\end{equation}
Here $\mathbf{1}[\cdot]$ is the indicator function, equal to one when its argument is true and zero otherwise. The average PSR is computed over all 480 test instances. Since all domains contain the same number of instances, the average score is equivalent to the macro-average over domains. Missing plans are counted as failures.

We additionally report goal progress (GP) as a supplementary measure of partial plan quality. For each test instance $i$, let $k_i$ denote the length of the maximal executable prefix of the generated plan $\hat{\pi}_i$, and let $s_i^{(k_i)}$ denote the state reached after executing this prefix from the initial state. Let $\mathcal{G}_i$ denote the set of goal atoms for instance $i$. The goal progress of $\hat{\pi}_i$ is defined as
\begin{equation}
	\operatorname{GP}(\hat{\pi}_i)
	=\frac{\left|\mathcal{G}_i \cap s_i^{(k_i)}\right|}{|\mathcal{G}_i|}.
	\label{eq:progress}
\end{equation}
This metric measures the fraction of goal conditions satisfied after executing the maximal valid prefix of the generated plan. Since full plan validity is determined using VAL, GP is used only as a supplementary metric to characterize partial progress alongside PSR. Detailed GP results are provided in Appendix F.5.

\subsection{Main Results}
\label{sec:main-results}

Table~\ref{tab:main} presents the problem-solving rates (PSRs) across eight planning domains. Overall, LLM-GRCR achieves the highest average PSR of 90.83\%, substantially outperforming both the strongest prompting baseline, Self-consistency CoT (45.83\%), and training baseline, Plan-SFT (76.25\%). LLM-GRCR also surpasses all evaluated external LLMs in average PSR, including Kimi-K2.5 (89.38\%). These results demonstrate the strong overall planning performance of LLM-GRCR across the evaluated domains.

The domain-wise results reveals that the performance improvements vary across planning environments. Compared with Plan-SFT, LLM-GRCR achieves particularly substantial gains on Blocksworld and Satellite, improving PSR by 50.00 and 35.00 percentage points, respectively. It also outperforms Plan-SFT on Briefcaseworld, Fridge, Grippers, and Logistics, matches its performance on Miconic, but performs slightly worse on Ferry. These observations indicate that the overall advantage of LLM-GRCR is primarily driven by substantial improvements in several domains rather than uniform gains across all domains.

Table~\ref{tab:main-difficulty} further investigates the planning performance across three difficulty levels. Compared with Plan-SFT, LLM-GRCR achieves a relatively modest improvement on easy instances (96.25\% vs. 92.50\%), whereas its advantage becomes increasingly pronounced as problem difficulty increases. Specifically, the performance gains reach 14.38 and 25.63 percentage points on medium and hard instances, respectively. These findings suggest that LLM-GRCR is particularly effective on more challenging planning problems, highlighting its ability to handle complex action dependencies.
\subsection{Ablation Study on Generation and Refinement}

Table~\ref{tab:component-ablation-v2} evaluates the contributions of key components in LLM-GRCR. Removing either GRPO optimization or critique-centric refinement reduces the average PSR from 90.83\% to 86.88\% and 81.67\%, respectively, while removing both further decreases it to 78.96\%. These results demonstrate that both components contribute to overall planning performance. Notably, removing refinement leads to a larger performance degradation than removing GRPO, highlighting the importance of identifying and repairing planning errors through iterative refinement.


The structured causal representation also plays an important role in plan generation. Without refinement, removing the causal representation reduces the average PSR from 81.67\% to 79.38\%. This suggests that explicitly modeling causal dependencies provides useful structural guidance for plan generation, contributing to improved planning performance.




\begin{figure}[t]
	\centering
	\includegraphics[width=0.7\linewidth]{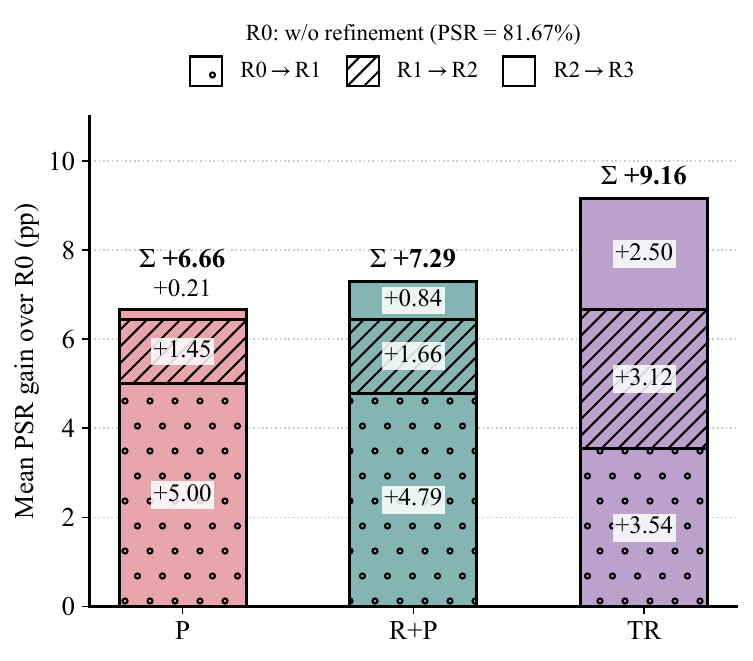}
	\caption{Cumulative average PSR gains relative to R0 across three refinement rounds under different revision strategies. }
	\label{fig:repair-format}
\end{figure}

\begin{figure}[t]
	\centering
	\includegraphics[width=0.65\linewidth]{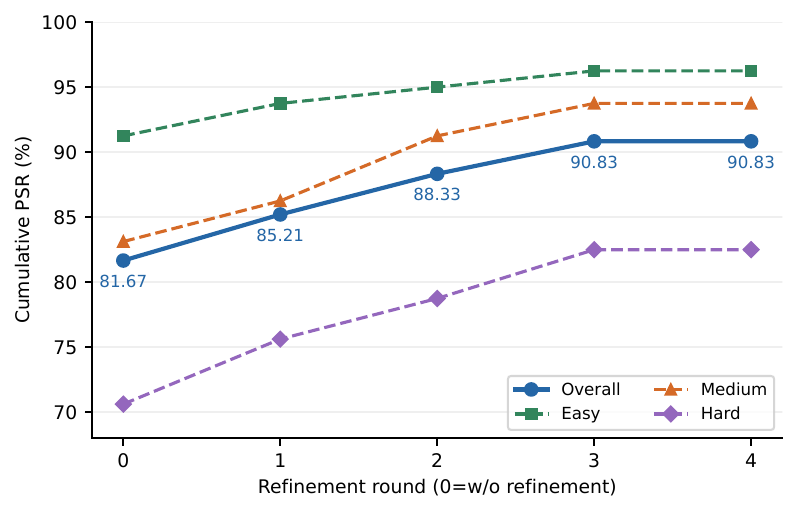}
	\caption{Effect of iterative targeted revision across successive rounds and difficulty levels.}
	\label{fig:refinement-rounds}
\end{figure}
\begin{figure}[t]
	\centering
	\includegraphics[width=0.7\linewidth]{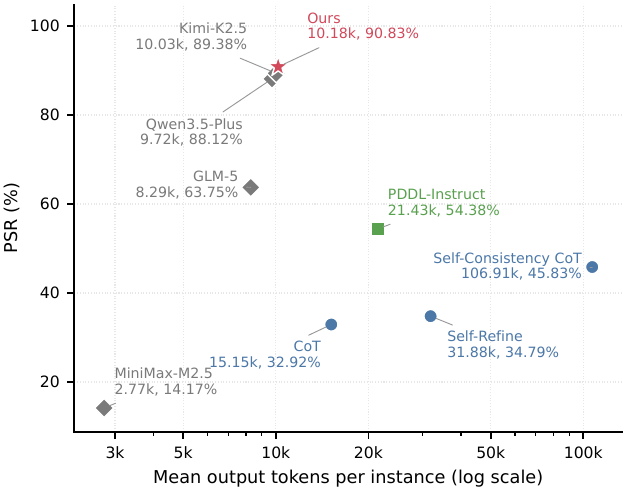}
	\caption{Planning performance versus token efficiency across representative LLM-based planning methods.}
	\label{fig:token-performance}
\end{figure}%

\subsection{Effect of Structural Reward}
\label{sec:reward-analysis}

Plan-validity rewards evaluate whether generated plans achieve the specified goals, but do not directly assess the causal dependencies underlying plan construction. To provide more explicit supervision for structured reasoning, LLM-GRCR introduces a structural reward \(R_{\mathrm{struct}}\) that encourages causal-link consistency, dependency coverage, and coherent goal updates. To evaluate its contribution, we compare models trained with and without \(R_{\mathrm{struct}}\) under the same refinement setting. As shown in Table~\ref{tab:reward-config}, incorporating the structural reward improves the average PSR from 89.37\% to 90.83\%, with notable gains on Grippers (86.67\% to 95.00\%), Satellite (90.00\% to 96.67\%), and Blocksworld (78.33\% to 83.33\%). These results demonstrate that structural supervision complements plan-level validity rewards by providing additional guidance on intermediate causal dependencies, contributing to overall planning performance.

%
\subsection{Effect of Revision Strategy}
\label{sec:repair-scope}

Figure~\ref{fig:repair-format} compares three revision strategies under the same initial generator. \emph{Reasoning+Plan Regeneration (R+P)} regenerates both the reasoning trace and the complete plan, \emph{Plan-Only Regeneration (P)} regenerates only the full action sequence, and \emph{Targeted Revision (TR)} performs localized revisions to the current action sequence based on the critique. All three configurations use 24,000 training examples and receive comparable feedback from the symbolic critique module. As shown in the figure, the regeneration-based strategies achieve larger gains in the first refinement round, but their improvements diminish rapidly in subsequent rounds. In contrast, Targeted Revision yields more sustained gains across rounds and achieves the largest cumulative improvement over the initial generation, reaching +9.16 percentage points by R3 (R3 denotes three refinement rounds), compared with +6.66 for P and +7.29 for R+P. These results suggest that targeted local revision is more effective at progressively improving an existing plan than repeatedly regenerating the full output.

%
\subsection{Effect of Iterative Refinement}
\label{sec:refinement-rounds}

A candidate plan may contain multiple interdependent errors that cannot be fully resolved through a single revision, motivating an iterative process of failure diagnosis and targeted revision. To evaluate the effectiveness of this process, Figure~\ref{fig:refinement-rounds}  illustrates how planning performance evolves across successive rounds of critique-centric refinement, with R0 denoting the initial generation without refinement. LLM-GRCR progressively improves the average PSR from 81.67\% at R0 to 85.21\%, 88.33\%, and 90.83\% after three refinement rounds. Notably, the improvements are more pronounced on hard instances, where initial plan generation is less reliable, while easy instances already achieve relatively high success rates before refinement. These results demonstrate that iterative critique-centric refinement effectively improves planning reliability by progressively correcting invalid plans, with particularly strong benefits for more challenging planning problems.

\subsection{Planning Performance and Token Efficiency}
\label{app:token-efficiency}

While explicit reasoning and iterative refinement improve planning reliability, their potential generation overhead raises concerns about token efficiency. To investigate this trade-off, Figure ~\ref{fig:token-performance} compares problem-solving rates (PSR) against mean output-token counts across reasoning-based planning approaches and LLM baselines. For a consistent comparison, all outputs are measured using the Qwen3.5-9B tokenizer, with token counts accumulated across candidate generations and subsequent refinement rounds. As shown in Figure~\ref{fig:token-performance}, LLM-GRCR achieves the highest PSR of 90.83\% on the 480-instance benchmark while generating approximately 10.18k output tokens per instance, fewer than the evaluated reasoning-based prompting methods. These results demonstrate a favorable balance between planning performance and output-token consumption, although input tokens and symbolic verification costs are not included in the comparison.

\section{Conclusion}



We presented LLM-GRCR, a framework that couples goal-directed plan generation with critique-centric refinement through a shared structured causal representation. The generator reasons backward from goal conditions to construct causal dependencies and candidate plans, while critique-centric refinement diagnoses planning failures and performs targeted revisions within the same structural framework. Experiments across eight planning domains show that LLM-GRCR achieves the highest average PSR among the evaluated baselines. Ablation studies further demonstrate the contributions of structured causal representation, GRPO optimization and targeted refinement. %
\bibliographystyle{ACM-Reference-Format}
\bibliography{sample}

\end{document}